\documentclass{article}
\usepackage{graphicx}
 \usepackage{amsmath} 
 \usepackage{multirow}
 \usepackage{multicol}
\usepackage{subcaption} 

 \usepackage[dblblindworkshop, final ]{neurips_2025}
\workshoptitle{Workshop on Space in Vision, Language, and Embodied AI}

\usepackage[utf8]{inputenc} 
\usepackage[T1]{fontenc}    
\usepackage{hyperref}       
\usepackage{url}            
\usepackage{booktabs}       
\usepackage{amsfonts}       
\usepackage{nicefrac}       
\usepackage{microtype}      
\usepackage{xcolor}         

\title{\textit{TriFusion-AE}: Language-Guided Depth and LiDAR Fusion for Robust Point Cloud Processing}

\author{%
    Susmit Neogi \\
  Department of Mechanical Engineering\\
  Indian Institute of Technology Bombay\\
  Mumbai, India \\
  \texttt{susmit.neogi@iitb.ac.in} \\
}

\begin{document}

\maketitle

\begin{abstract}
  LiDAR-based perception is central to autonomous driving and robotics, yet raw point clouds remain highly vulnerable to noise, occlusion, and adversarial corruptions. Autoencoders offer a natural framework for denoising and reconstruction, but their performance degrades under challenging real-world conditions. In this work, we propose \textit{TriFusion-AE}, a multimodal cross-attention autoencoder that integrates textual priors, monocular depth maps from multi-view images, and LiDAR point clouds to improve robustness. By aligning semantic cues from text, geometric (depth) features from images, and spatial structure from LiDAR, \textit{TriFusion-AE} learns representations that are resilient to stochastic noise and adversarial perturbations. Interestingly, while showing limited gains under mild perturbations, our model achieves significantly more robust reconstruction under strong adversarial attacks and heavy noise, where CNN-based autoencoders collapse. We evaluate on the nuScenes-mini dataset to reflect realistic low-data deployment scenarios. Our multimodal fusion framework is designed to be model-agnostic, enabling seamless integration with any CNN-based point cloud autoencoder for joint representation learning.
\end{abstract}

\section{Introduction}
Autonomous vehicles and mobile robots increasingly rely on LiDAR sensors for reliable perception and navigation. LiDAR offers rich 3D structural information about the environment, but raw point clouds are notoriously fragile: even small perturbations from sensor noise, occlusion, or targeted adversarial attacks can lead to drastic failures in downstream tasks such as detection and mapping [1]. The need for robust point cloud representations is therefore important for safety-critical applications.

Autoencoders (AEs) have emerged as a natural solution for denoising and reconstructing corrupted LiDAR data, serving as a backbone for both perception and robustness pipelines. However, single-modality AEs often collapse under challenging conditions, particularly in scenarios with heavy noise or adversarial perturbations. In practice, autonomous systems rarely rely on LiDAR alone; multi-sensory information from cameras and textual priors (e.g., high-level scene descriptions or captions) can provide complementary semantic and contextual cues. Leveraging this multi-modality remains underexplored in the context of robust LiDAR reconstruction.

In this work, we propose \textit{TriFusion-AE}, a multimodal fusion autoencoder that integrates LiDAR point clouds, monocular depth maps from multi-view images, and textual scene descriptions through weighted cross-attention mechanisms. By aligning semantic priors from text, geometric cues from images, and spatial structure from LiDAR, \textit{TriFusion-AE} aims to enhance reconstruction robustness against both stochastic noise and adversarial attacks. Interestingly, our results suggest that while \textit{TriFusion-AE} provides limited improvements under weak perturbations, it demonstrates significant gains under strong corruptions, where CNN-based AEs tend to collapse. This shows strong adversarial robustness of multimodal fusion against unimodal counterparts. 

We evaluate our framework on the nuScenes-mini [2] dataset, a small-scale realistic benchmark, to mimic resource-constrained training regimes. We find that while attention-based fusion models generally require large datasets to outperform convolutional baselines, \textit{TriFusion-AE} nonetheless shows meaningful robustness advantages in adversarially challenging conditions. This highlights the potential of multimodal fusion for LiDAR robustness, even in low-data regimes, and motivates future exploration at larger scales.

\section{Related Works}

Multimodal fusion has become fundamental in autonomous perception, driven by the complementary nature of LiDAR and camera data. Early approaches often relied on concatenation or simple projection methods, such as point-level fusion augmented with camera features, but suffered from semantic degradation and inefficient alignment [3]. Alternately, DeepFusion introduces learnable geometric alignment mechanisms to better integrate deep LiDAR and image features, achieving strong performance and generalization [4]. More recently, SparseFusion reformulates fusion in terms of sparse candidate representations, achieving both efficiency and accuracy through attention-based candidate merging [5]. Complementing these, DepthFusion proposes a depth-aware hybrid feature fusion mechanism that adaptively weights point cloud and image modalities at both global and local levels via depth encoding, resulting in improved 3D object detection [6].

In parallel, research has begun extending sensor fusion into generative and world-modeling domains. For example, MUVO investigates fusion strategies within a multimodal world model, emphasizing non-BEV fusion and the prediction of 3D occupancy for future frames [7]. On the autoencoder front, masked autoencoder variants like UniM2AE aim to jointly reconstruct multimodal inputs, encouraging unified representations across modalities and hinting at more general-purpose fusion encoders [8].

Finally, recent advancements in vision-language fusion highlight the growing importance of integrating textual information into spatial understanding. VLM-E2E, for instance, uses a BEV-Text learnable weighting scheme to dynamically balance visual and textual features allowing the model to adaptively shift attention based on context [9].

Our work, \textit{TriFusion-AE}, builds upon these foundations by introducing an autoencoder architecture that jointly fuses LiDAR, multi-view monocular depth, and textual modalities. We further explore how stochastic, FGSM, and PGD adversarial perturbations interact with multimodal reconstructions, assessing whether joint encoding provide better resilience against noise perturbations, compared to unimodal counter parts.

\section{Methodology}

\begin{figure*}[t]
\centering
\includegraphics[width=1\columnwidth]{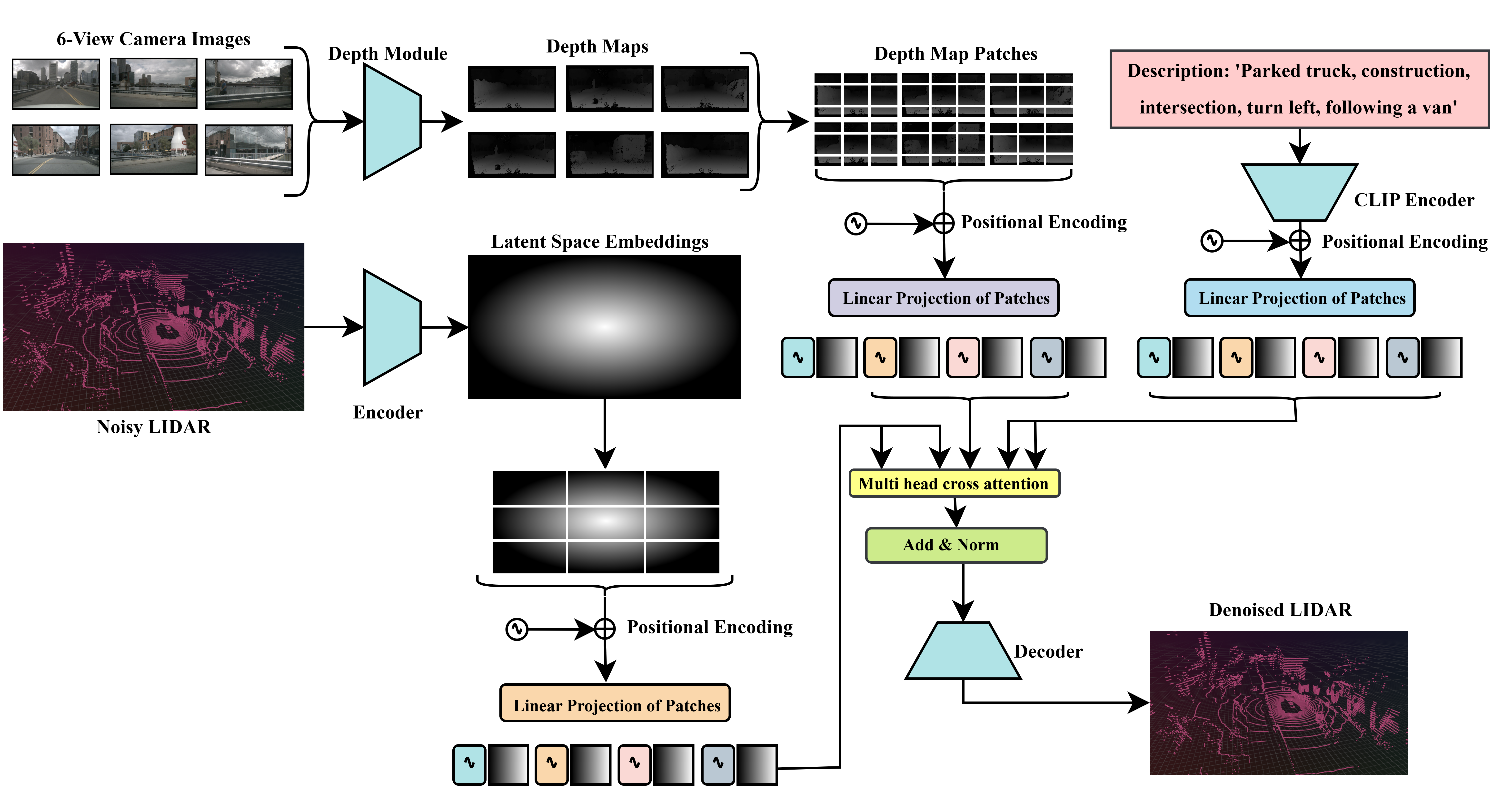} 
\caption{\textit{TriFusion-AE}: Model Architecture
}
\label{fig1}
\end{figure*}

In this section, we describe our proposed \textit{TriFusion-AE} framework for adversarially robust point cloud reconstruction. The method integrates three complementary modalities -- LiDAR point clouds, depth map from multi-view  images, and textual scene descriptions via cross-attention fusion to obtain semantically grounded and spatially consistent reconstructions.  

\subsection{Problem Formulation}

Let $x^* \in \mathbb{R}^{N \times d}$ denote a clean input LiDAR point cloud with $N$ points in $d$ dimensions (typically $d=3$ or $d=4$ including intensity). The corrupted input $x$ is generated under a perturbation model:
\begin{equation}
    x = x^* + \delta, \quad \|\delta\|_{p} \leq \epsilon,
\end{equation}
where $\delta$ denotes adversarial or stochastic noise bounded by an $L_p$ norm. The objective of the autoencoder is to reconstruct a clean approximation $\hat{x}$ minimizing a reconstruction loss:
\begin{equation}
    \mathcal{L}_{\text{rec}} = \mathcal{D}(x^*, \hat{x}),
\end{equation}
where $\mathcal{D}(\cdot)$ is a similarity measure such as Mean Squared Error (MSE).

\subsection{Multimodal Encoding}

We consider three modalities as input streams: LiDAR point cloud, multi-view RGB images and text captions of the scene.

\begin{figure*}[h!]
    \centering
    \begin{subfigure}[t]{0.48\textwidth}
        \centering
        \includegraphics[width=\linewidth]{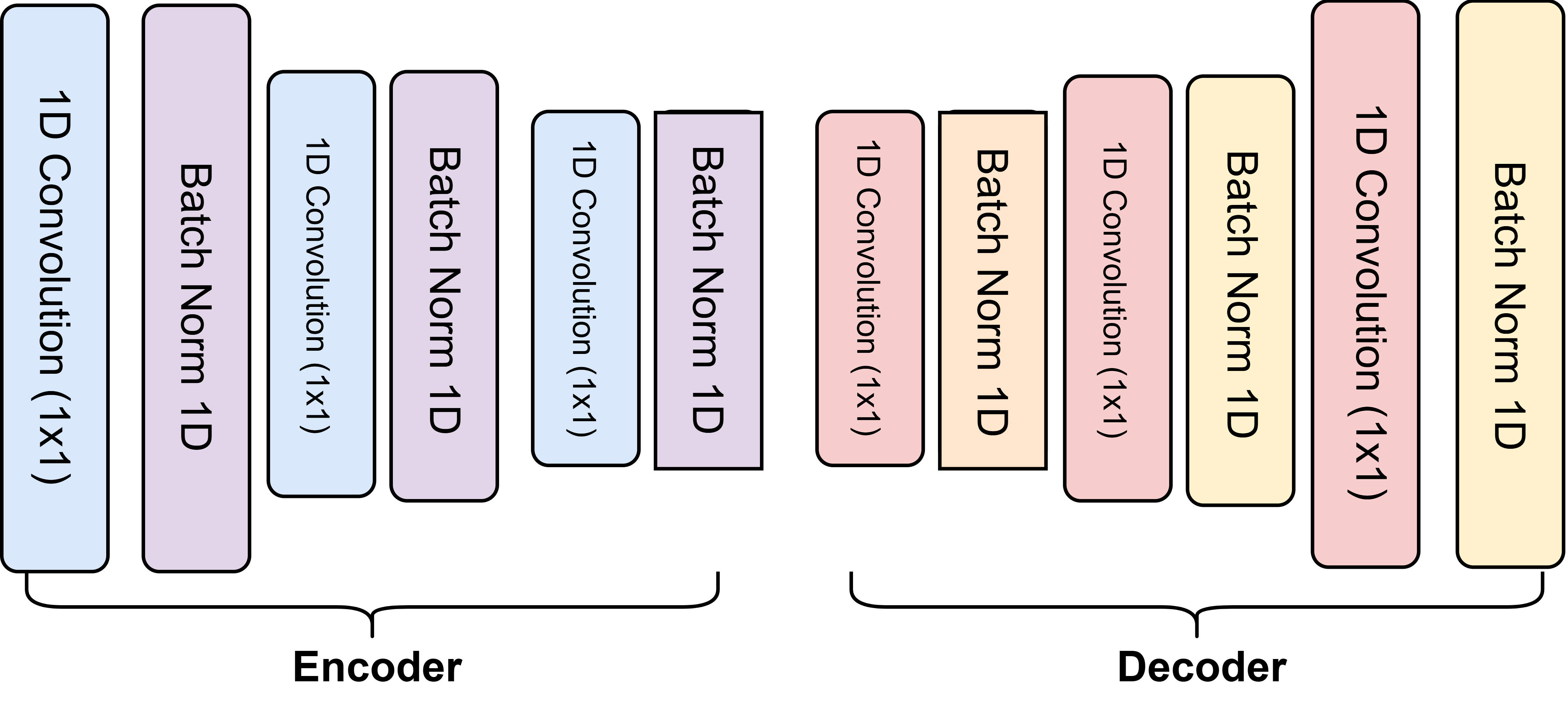}
        \caption{}
        \label{fig:sub-a}
    \end{subfigure}
    \hfill
    \begin{subfigure}[t]{0.48\textwidth}
        \centering
        \includegraphics[width=\linewidth]{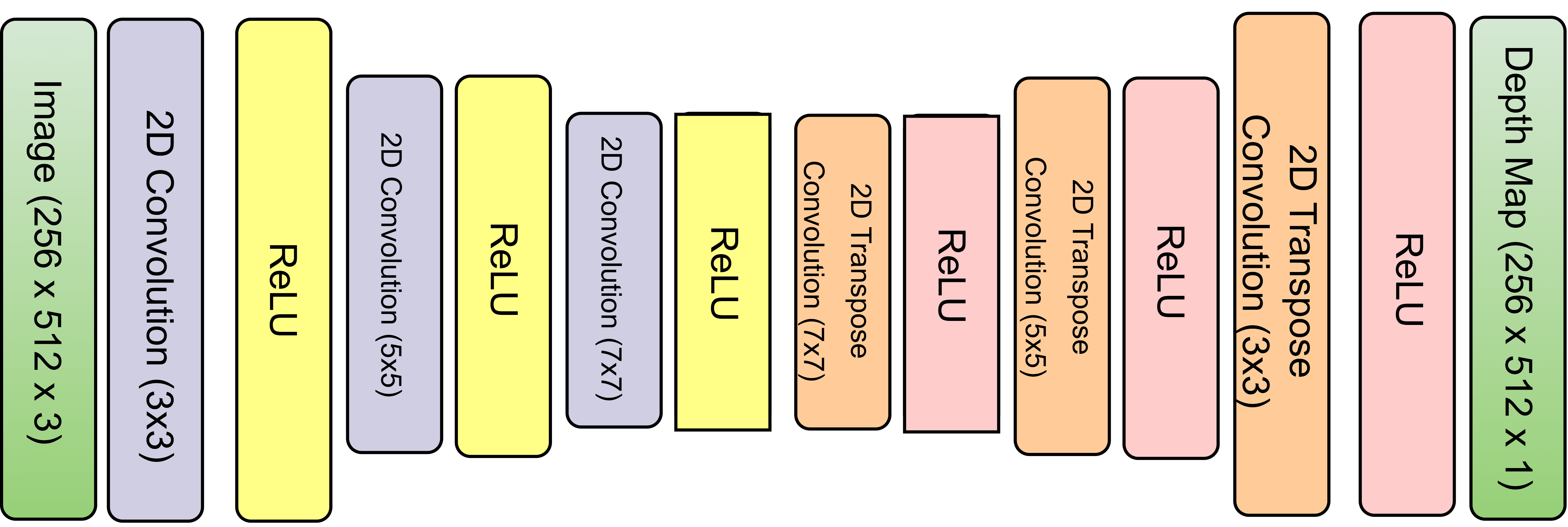}
        \caption{}
        \label{fig:sub-b}
    \end{subfigure}
    
    \caption{(a) CNN Autoencoder architecture (b) Depth Estimation module architecture}
    \label{fig:two-subfigs}
\end{figure*}

For encoding the LiDAR point cloud (Figure~\ref{fig1}), we employ a low-dimensional latent space autoencoder with a single dimensional bottleneck layer (Figure~ \ref{fig:sub-a}). This latent feature map is further divided into $1 \times 1$ patches and projected to a common embedding dimension $d = 256$. We call the LiDAR latent features as $z_L$. The range of values in LiDAR point cloud data is $[-255,255]$. The $4$ input channels of the point cloud are compressed into a single latent channel, effectively aligning the representation with the dimensionality of monocular depth maps.  This dimensionality reduction not only enforces a compact embedding that captures the most salient geometric structures, but also facilitates cross-modal fusion with camera-based depth representations. Moreover, the bottleneck constraint acts as a form of regularization, encouraging robustness against adversarial perturbations and noise by preventing the network from overfitting to high-dimensional artifacts.

We design a CNN based monocular depth estimation model (Figure~\ref{fig:sub-b}), which is essentially another encoder-decoder model that takes RGB images of shape $(X,Y,3)$ and returns a depth map of shape $(X,Y,1)$. We use a relatively light weight depth module as a regularization technique to prevent overfitting. It is further divided into $16 \times 16$ patches and projected to the same embedding dimension $d$. Describing the depth information of the entire scene, like LiDAR, requires a panoramic view, thus requiring stitching of the multi-view depth maps. However, our model takes care of this spatial relationship using positional embeddings of every patch of the depth maps. We call this depth feature map as $z_D$.

For textual captions of the scene, we use standard ViT [15] based frozen CLIP [14] weights, which encode the text into a 512-dimensional embedding space. This is further divided into $1 \times 1$ patches and projected to the same embedding dimension $d$. We call this text embedding as $z_T$.

Thus, the joint representation before fusion is
\begin{equation}
    \mathbf{Z} = \{\mathbf{z}_{L}, \mathbf{z}_{D}, \mathbf{z}_{T}\}.
\end{equation}


\subsection{Cross-Attention Fusion}

To align features across modalities, we introduce multi-head cross-attention module inspired by transformer architecture [10].  Let, $\{L, D, T\}$ be the modalities of LiDAR, depth and text respectively. For a single attention head, we have the query matrix $Q_L$ corresponding to LiDAR, and the key and value matrices $K_D$, $V_D$ for depth and $K_T$, $V_T$ for text respectively. So, we have two attention terms:
\begin{equation}
    \text{Attn}(Q_{L}, K_{D}, V_{D}) = \text{softmax}\left( \frac{Q_{L}K_{D}^\top}{\sqrt{d}} \right) V_{D},
\end{equation}
\begin{equation}
   \text{Attn}(Q_{L}, K_{T}, V_{T}) = \text{softmax}\left( \frac{Q_{L}K_{T}^\top}{\sqrt{d}} \right) V_{T},
\end{equation}

The fused representation for the enriched LiDAR embedding $z_L^*$ is obtained by a weighted average:
\begin{equation}
    \mathbf{z}_{L}^{*} = \lambda_1 \cdot \text{Attn}(Q_{L}, K_{D}, V_{D}) + \lambda_2 \cdot \text{Attn}(Q_{L}, K_{T}, V_{T})
\end{equation}

In our architecture, we have equal weights $\lambda_1 = \lambda_2 = 0.5$

\subsection{Decoder and Reconstruction}

To reconstruct the denoised LiDAR point cloud from the enriched latent embedding $z_L^*$, we have a two stage decoder. First decoder is a linear mapping, transforming the feature map from the embedding dimension, back to latent space dimension. The following decoder is a CNN based model, upsampling the single channeled latent feature map back to the original 4 channels of LiDAR point cloud data.





\subsection{Mathematical Model}

Let the clean point cloud be $x^* \in \mathbb{R}^d$. The LiDAR observation is modeled as
\begin{equation}
    x = x^* + n_L + a ,
\end{equation}
where $n_L$ is stochastic noise with $\mathbb{E}[n_L] = 0$, $\mathrm{Var}(n_L) = \sigma_L^2 I$, and $a$ is an adversarial perturbation bounded as $\|a\|_\infty \leq \epsilon$.  

The image--text prior produces a data-dependent estimate $g(I,T)$ of $x$. In expectation, we write
\begin{equation}
    g = x^* + b + n_P ,
\end{equation}
where $b$ is a bias due to imperfect priors or limited data, and $\mathrm{Var}(n_P) = \sigma_P^2 I$.  

Let us assume a fusion estimator as an abstraction of local cross-attention weighting:
\begin{equation}
    \hat{x}(\alpha) = \alpha x + (1-\alpha) g, \quad \alpha \in [0,1].
\end{equation}
Here $\alpha$ represents the weight placed on LiDAR versus prior information. Assume $n_L$ and $n_P$ are uncorrelated with each other and with $x^*$.  
\begin{align}
    \mathrm{Error}(\alpha) 
    &= \mathbb{E}\|\hat{x}(\alpha) - x^*\|_2^2 \\
    &= \alpha^2 \, \mathbb{E}\|n_L + a\|_2^2 \;+\; (1-\alpha)^2 \, \mathbb{E}\|b + n_P\|_2^2 .
\end{align}

Define the effective errors for LiDAR $(x)$ as $V_x$ and for text-image prior $(P)$ as $V_P$:
\begin{equation}
    V_x := \mathbb{E}\|n_L + a\|_2^2 = \sigma_L^2 + {\|a\|_2^2}, \qquad 
    V_P :=\mathbb{E}\|b + n_P\|_2^2 =  \sigma_P^2 + {\|b\|_2^2}.
\end{equation}
\begin{equation}
    \mathrm{Error}(\alpha) = \alpha^2 V_x + (1-\alpha)^2 V_P.
\end{equation}

The minimizing weight, which is eventually achieved by optimization algorithm (Adam, in our case) is:
\begin{equation}
    \alpha^\star = \frac{V_P}{V_x + V_P}, 
    \qquad 
    \mathrm{Error}_{\min} = \frac{V_x V_P}{V_x + V_P}.
\end{equation}

Comparing with the LiDAR-only risk, $\mathrm{Error}_x = V_x$, fusion theoretically yields improvement, as the following always holds true: 
\begin{equation}
  \mathrm{ Error}_{min} = \frac{V_x V_P}{V_x + V_P} < V_x = \mathrm{Error}_x
\end{equation}

\section{Experiments}

We experiment with three kinds of noise perturbation: (1) Stochastic (Gaussian) Noise, (2) Fast Gradient Sign Method (FGSM) adversarial attack, (3) Projected Gradient Descent Method (PGD) adversarial attack. 
We do not perform adversarial training; instead, we test the model's inherent (natural) robustness.
For fair comparison, the encoder and decoder architectures in \textit{TriFusion-AE} are the same as those used in the CNN autoencoder baseline. We report the reconstruction MSE Loss, PSNR (in dB) and the marginal change in PSNR and MSE for TriFusion-AE compared to CNN AE.
We train both the CNN AE and \textit{TriFusion-AE} model on nuScenes-mini dataset with train-test split of 0.8-0.2 (seed=$32$), on RTX 3050 8GB GPU. 
\subsection{Stochastic Noise}

We begin by adding a Gaussian noise $\mathcal{N}(0,1)$ to the LiDAR data. Progressively, we increase the noise intensity, by multiplying by a scalar intensity factor $\alpha$, essentially scaling the peak noise intensity by $\alpha$ and noise variance by $\alpha^2$.  As mentioned previously, this stochastic noise is a $L_2$ norm bounded perturbation. 

We report the MSE and PSNR of the reconstructed denoised LiDAR and the original LiDAR point cloud, using both CNN based autoencoder and \textit{TriFusion-AE}, in table~\ref{t1}, against different $\alpha$ values. 



\begin{table}[h!]
\centering
\small
\caption{Comparison of CNN AE and \textit{TriFusion-AE} across different noise levels ($\alpha$)}
\label{t1}
\renewcommand{\arraystretch}{1.3} 
\setlength{\tabcolsep}{8pt}       
\begin{tabular}{|c|c|c|c|c|c|c|c|}
\hline
\multirow{2}{*}{\textbf{S.No}} & \multirow{2}{*}{$\mathbf{\alpha}$} & \multicolumn{3}{c|}{\textbf{MSE} $\mathbf{(\downarrow)}$} & \multicolumn{3}{c|}{\textbf{PSNR (dB)}$\mathbf{(\uparrow)}$} \\
\cline{3-8}
 & & \textbf{CNN AE} & \textit{\textbf{TriFusion-AE}} & \textbf{\%$\Delta$} & \textbf{CNN AE} & \textit{\textbf{TriFusion-AE}} & \textbf{\%$\Delta$} \\
\hline
1 & 1  & 155.779 & 205.711 & +32.06\% & 23.212 & 21.9861 & -5.28\% \\
2 & 2  & 159.970 & 208.783 & +30.52\% & 23.164 & 22.0806 & -4.68\% \\
3 & 5  & 187.345 & 229.624 & +22.56\% & 22.952 & 22.0799 & -3.80\% \\
4 & 10 & 278.368 & 303.009 & +8.85\%  & 22.245 & 21.7089 & -2.41\% \\
\textbf{5} & \textbf{20} & \textbf{626.165} & \textbf{595.502} & \textbf{-4.90\%}  & \textbf{20.019} & \textbf{20.3931} & \textbf{+1.87\%} \\
6 & 30 & 1188.029 & 1086.648 & -8.55\%  & 19.332 & 19.9335 & +3.11\% \\
7 & 40 & 1953.229 & 1785.144 & -8.61\%  & 19.466 & 19.5895 & +0.64\% \\
8 & 50 & 2913.211 & 2666.089 & -8.48\%  & 19.006 & 18.8773 & -0.68\% \\
9 & 60 & 4080.566 & 3762.040 & -7.80\%  & 18.295 & 18.5610 & +1.46\% \\
\hline
\end{tabular}
\end{table}

\begin{figure*}[t]
\centering
\includegraphics[width=1\columnwidth]{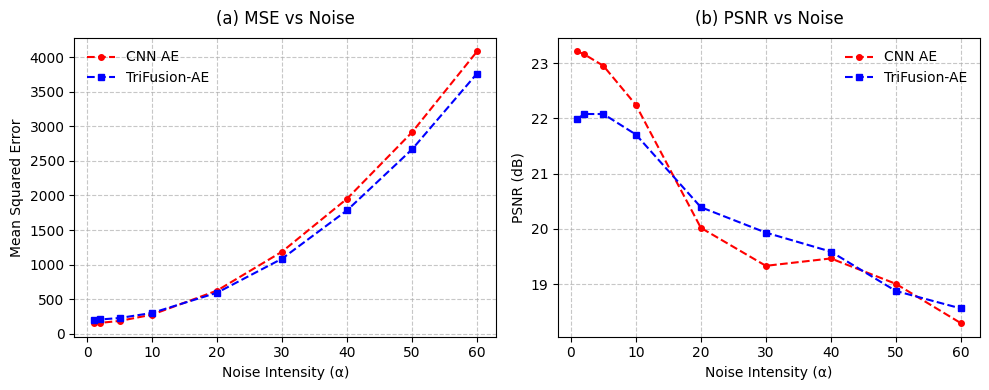} 
\caption{MSE and PSNR Trends with Stochastic Noise Intensity ($\alpha$) for CNN AE and \textit{TriFusion-AE}}
\label{fig2}
\end{figure*}

\paragraph{Analysis} 
From Figure~\ref{fig2}, we observe that the Mean Squared Error (MSE) increases monotonically with noise intensity for both models, which is expected due to amplified signal corruption. However, \textit{TriFusion-AE} consistently achieves lower MSE compared to CNN-AE, particularly at higher noise levels (e.g., $\alpha \geq 40$), indicating stronger resilience against perturbations. 
We also observe the Peak Signal-to-Noise Ratio (PSNR), where a higher value reflects better reconstruction quality. Both models exhibit decreasing PSNR with increasing noise, yet \textit{TriFusion-AE} maintains a noticeable margin over CNN-AE in the moderate-to-high noise regime. Interestingly, while CNN-AE undergoes a sharper drop in PSNR beyond $\epsilon=20$, \textit{TriFusion-AE} degrades more gradually.

\subsection{FGSM}
To evaluate the robustness of our model against adversarial perturbations, we employed the Fast Gradient Sign Method (FGSM) [11]. FGSM generates adversarial examples by perturbing the original input $x$ in the direction of the gradient of the loss with respect to the input. Formally, the adversarial input is constructed as:
\begin{equation}
    x^{\text{adv}} = x + \epsilon \cdot \text{sign}\!\left( \nabla_{x} J(\theta, x, y) \right)
\end{equation}
where $J(\theta, x, y)$ denotes the loss function of the model parameterized by $\theta$ for input $x$ and true label $y$, and $\epsilon$ is the perturbation budget that controls the $\ell_\infty$ norm of the adversarial noise. Intuitively, this perturbation pushes the input in the direction that maximally increases the model's loss, while keeping the modification visually negligible.
We choose the autoencoder model as the parametrized model $\theta$. In our setting, we modify the classification label term as denoised/actual $x^*$:
\begin{equation}
    x^{\text{adv}} = x + \epsilon \cdot \text{sign}\!\left( \nabla_{x} J(\theta, x, x^*) \right)
\end{equation}

We use Cleverhans [12] to perform the adversarial attacks. We report the MSE and PSNR of the reconstructed denoised LiDAR and the original LiDAR point cloud, using both CNN based autoencoder and \textit{TriFusion-AE}, in table~\ref{t2}, against different $\epsilon$ values. 
Note that the maximum value of LiDAR point cloud is 255, so $\epsilon$ of 70 is $\approx 28\%$ noise, which is a strong but reasonable perturbation. Also, a higher $\epsilon$ means a longer step in gradient ascent, not necessarily meaning a stronger attack.


\begin{table}[h!]
\centering
\small
\caption{Comparison of CNN AE and \textit{TriFusion-AE} under FGSM attack with different $\epsilon$ values.}
\label{t2}
\renewcommand{\arraystretch}{1.3} 
\setlength{\tabcolsep}{8pt}       
\begin{tabular}{|c|c|c|c|c|c|c|c|}
\hline
\multirow{2}{*}{\textbf{S.No}} & \multirow{2}{*}{\boldmath{$\epsilon$}} & \multicolumn{3}{c|}{\textbf{MSE}$\mathbf{(\downarrow)}$} & \multicolumn{3}{c|}{\textbf{PSNR (dB)}$\mathbf{(\uparrow)}$} \\
\cline{3-8}
 & & \textbf{CNN AE} & \textit{\textbf{TriFusion-AE}} & \textbf{\%$\Delta$} & \textbf{CNN AE} & \textit{\textbf{TriFusion-AE}} & \textbf{\%$\Delta$} \\
\hline
1  & 0.1  & 154.555 & 204.750 & +32.46\%  & 23.143 & 21.922 & -5.29\% \\
2  & 0.5  & 156.179 & 204.425 & +30.91\%  & 23.098 & 21.929 & -5.06\% \\
3  & 1.0  & 153.792 & 204.898 & +33.21\%  & 23.165 & 21.919 & -5.38\% \\
4  & 5.0  & 163.384 & 204.414 & +25.14\%  & 22.902 & 21.929 & -4.25\% \\
\textbf{5}  & \textbf{10.0} & \textbf{379.048} & \textbf{206.895} & \textbf{-45.40\%}  & \textbf{19.247} & \textbf{21.877} & \textbf{+13.66\%} \\
6  & 20.0 & 276.109 & 219.741 & -20.39\%  & 20.623 & 21.615 & +4.81\% \\
7  & 40.0 & 271.589 & 227.402 & -16.27\%  & 20.695 & 21.466 & +3.73\% \\
8  & 55.0 & 985.440 & 283.270 & -71.26\%  & 15.098 & 20.512 & +35.84\% \\
9  & 60.0 & 3710.259 & 279.117 & -92.48\%  & 9.340  & 20.576 & +120.26\% \\
10 & 70.0 & 1095.941 & 278.323 & -74.61\%  & 14.636 & 20.589 & +40.69\% \\
\hline
\end{tabular}
\end{table}

\begin{figure*}[t]
\centering
\includegraphics[width=1\columnwidth]{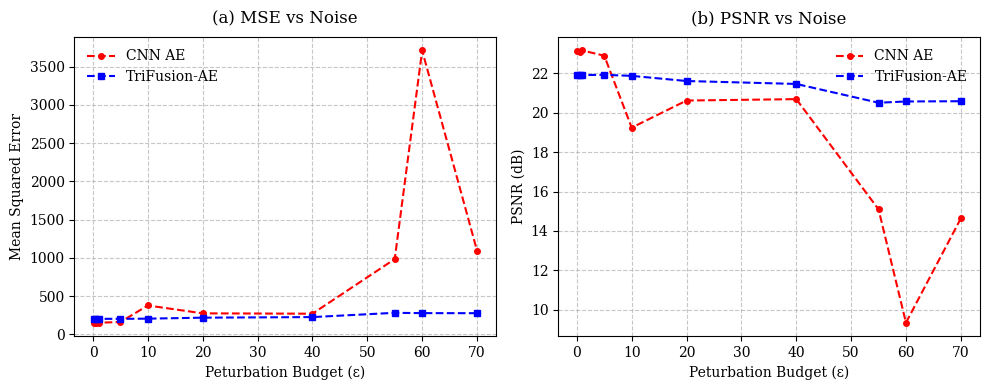} 
\caption{MSE and PSNR Trends with FGSM perturbation budget ($\epsilon$) for CNN AE and \textit{TriFusion-AE}
}
\label{fig3}
\end{figure*}

\paragraph{Analysis}
As shown in Figure~\ref{fig3}, the mean squared error (MSE) of the CNN AE model increases sharply when the perturbation budget exceeds $\epsilon=50$, reaching values above $1500$, whereas \textit{TriFusion-AE} maintains consistently low error across all noise levels. A similar trend is observed in the PSNR results: while the CNN model undergoes a steep degradation beyond $\epsilon=40$, dropping below $10$ dB at $\epsilon=70$, \textit{TriFusion-AE} sustains stable reconstruction quality with PSNR values above $20$ dB throughout. These results highlight the superior resilience of \textit{TriFusion-AE} to noise perturbations, demonstrating its capability to preserve reconstruction fidelity under increasingly challenging conditions.

\subsection{PGD}
We further consider the \textbf{Projected Gradient Descent Method (PGD)} attack, which is widely regarded as a universal first-order adversary [13]. Given a classifier $f_\theta(\cdot)$ with parameters $\theta$, input $x \in \mathbb{R}^d$, and true label $y$, the PGD adversarial example is obtained via iterative gradient-based perturbations constrained within an $\ell_\infty$-ball of radius $\epsilon$. Starting from an initial perturbation $x^{(0)} = x + \delta$, where $\delta \sim \mathcal{U}(-\epsilon, \epsilon)$, the adversarial input is updated as:
\begin{equation}
x^{(t+1)} = \Pi_{\mathcal{B}_\epsilon(x)} \Big( x^{(t)} + \alpha \cdot \text{sign}\big( \nabla_{x^{(t)}} \mathcal{L}(f_\theta(x^{(t)}), y) \big) \Big),
\end{equation}
where $\alpha$ is the step size, $\mathcal{L}$ is the training loss, and $\Pi_{\mathcal{B}_\epsilon(x)}(\cdot)$ denotes projection onto the $\ell_\infty$-ball $\mathcal{B}_\epsilon(x) = \{x' : \|x'-x\|_\infty \leq \epsilon\}$. Unlike the single-step FGSM method, PGD applies multiple iterations ($T$) of small updates, allowing it to explore the local neighborhood more effectively and often producing stronger adversarial examples. Similar to the previous setting, we adapt the PGD iterations for autoencoder reconstruction (replacing classification labels with the clean signal $x^*$):
\begin{equation}
x^{(t+1)} = \Pi_{\mathcal{B}_\epsilon(x)} \Big( x^{(t)} + \alpha \cdot \text{sign}\big( \nabla_{x^{(t)}} \mathcal{L}(f_\theta(x^{(t)}), x^*) \big) \Big),
\end{equation}
where, $\mathcal{L}$ is MSE Loss and $x^*$ is the original point cloud vector.
We keep the step size of each attack iteration as $\epsilon_{iter} = 0.01$ and number of iterations $n_{iter}=40$. We report the MSE and PSNR of the reconstructed denoised LiDAR and the original LiDAR point cloud, using both CNN based autoencoder and \textit{TriFusion-AE}, in table~\ref{t3}, against different $\alpha$ values. 



\begin{table}[h!]
\centering
\small
\caption{Comparison of CNN AE and \textit{TriFusion-AE} under PGD attack with different $\alpha$ values.}
\label{t3}
\renewcommand{\arraystretch}{1.3} 
\setlength{\tabcolsep}{8pt}       
\begin{tabular}{|c|c|c|c|c|c|c|c|}
\hline
\multirow{2}{*}{\textbf{S.No}} & \multirow{2}{*}{\boldmath{$\alpha$}} & \multicolumn{3}{c|}{\textbf{MSE}$\mathbf{(\downarrow)}$} & \multicolumn{3}{c|}{\textbf{PSNR (dB)}$\mathbf{(\uparrow)}$} \\
\cline{3-8}
 & & \textbf{CNN AE} & \textit{\textbf{TriFusion-AE}} & \textbf{\%$\Delta$} & \textbf{CNN AE} & \textit{\textbf{TriFusion-AE}} & \textbf{\%$\Delta$} \\
\hline
1  & 0.1  & 155.01 & 204.74 & +32.07\%  & 23.13 & 21.92 & -5.09\% \\
2  & 0.5  & 155.03 & 204.84 & +32.10\%  & 23.13 & 21.92 & -5.09\% \\
3  & 1.0  & 154.64 & 204.75 & +32.36\%  & 23.14 & 21.92 & -5.29\% \\
4  & 5.0  & 158.70 & 204.96 & +29.14\%  & 23.03 & 21.92 & -4.81\% \\
5  & 10.0 & 171.76 & 205.72 & +19.78\%  & 22.68 & 21.90 & -3.44\% \\
\textbf{6}  & \textbf{20.0} & \textbf{213.14} & \textbf{207.94} & \textbf{-2.44\%}   & \textbf{21.75} & \textbf{21.85} & \textbf{ +0.46\%} \\
7  & 40.0 & 345.94 & 220.49 & -36.26\%  & 19.64 & 21.60 & +9.98\% \\
8  & 55.0 & 490.86 & 233.89 & -52.35\%  & 18.12 & 21.34 & +17.77\% \\
9  & 60.0 & 548.73 & 237.93 & -56.65\%  & 17.64 & 21.27 & +20.54\% \\
10 & 70.0 & 659.05 & 244.73 & -62.86\%  & 16.84 & 21.15 & +25.60\% \\
\hline
\end{tabular}
\end{table}

\begin{figure*}[t]
\centering
\includegraphics[width=1\columnwidth]{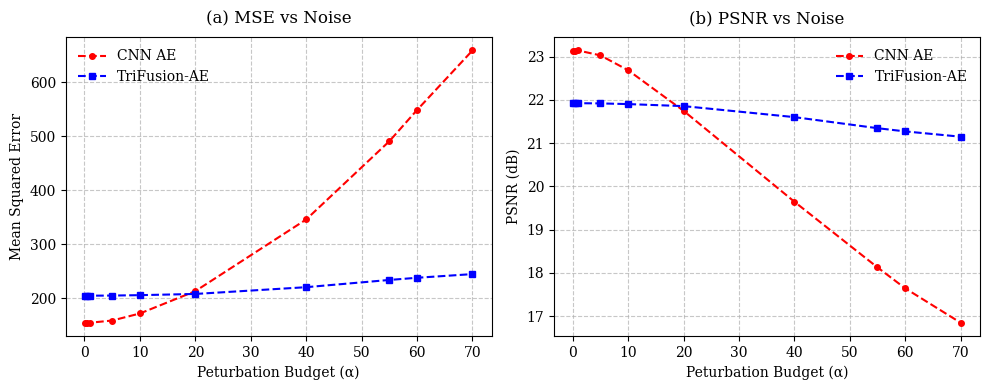} 
\caption{MSE and PSNR Trends with PGD perturbation budget ($\alpha$) for CNN AE and \textit{TriFusion-AE}
}
\label{fig4}
\end{figure*}

\paragraph{Analysis}
Figure~\ref{fig4} illustrates the robustness comparison of CNN-AE and \textit{TriFusion-AE} against adversarial perturbations crafted using the PGD attack. As shown in the figure, the Mean Squared Error (MSE) of CNN AE grows sharply with the perturbation budget $\alpha$, exceeding $600$ at higher perturbation levels, while \textit{TriFusion-AE} exhibits significantly lower error and a more stable trend across the entire range. 
Similarly, Figure~\ref{fig4} also reports the Peak Signal-to-Noise Ratio (PSNR) variation under noise. The PSNR of CNN AE rapidly declines below $18$ dB as $\alpha$ increases, whereas \textit{TriFusion-AE} maintains relatively higher PSNR, consistently above $20$ dB even at large perturbations. 

\section {Sensitivity}
 A reliable metric for robustness is sensitivity towards adversarial attacks and stochastic noise. We fit a linear regression model as: 
 \\
 \begin{equation}
Y = m \cdot X + C + \nu
 \end{equation}

 where, $Y$ is PSNR and MSE, while $X$ is the dependent variable (noise intensity $\alpha$ for stochastic noise, perturbation budget $\epsilon$ for FGSM and $\alpha$ for PGD attack.), $m$ is the slope, $C$ is the bias, $\nu \sim \mathcal{N}(0, 1)$ is the gaussian error term. 
Sensitivity $(S)$ can be calculated as :
 \begin{equation}
     S = |m| = \left| \frac{\partial Y}{\partial X} \right|
 \end{equation}
We report the CNN AE and \textit{TriFusion-AE} sensitivity ($S_{MSE}$ for MSE and $S_{PSNR}$ for PSNR) for stochastic noise, FGSM and PGD perturbations in table \ref{tab-sens}.

\begin{table}[h!]
\centering
\caption{Sensitivity of CNN AE and \textit{TriFusion-AE} under different perturbations.}
\label{tab-sens}
\renewcommand{\arraystretch}{1.2} 
\setlength{\tabcolsep}{8pt}       
\begin{tabular}{|c|c|c|c|}
\hline
\textbf{Perturbation} & \textbf{Model} & \textbf{$\mathbf{S_{MSE} (\downarrow)}$} & \textbf{$\mathbf{{S_{PSNR} (\downarrow)}}$} \\
\hline
\multirow{2}{*}{Stochastic Noise} 
& CNN AE       & 62.70 & 0.08  \\
\cline{2-4}
& \textit{TriFusion-AE} & \textbf{56.39} & \textbf{0.06}  \\
\hline
\multirow{2}{*}{FGSM} 
& CNN AE       & 27.39 & 0.15  \\
\cline{2-4}
& \textit{TriFusion-AE} &\textbf{ 1.23}  & \textbf{0.022} \\
\hline
\multirow{2}{*}{PGD} 
& CNN AE       & 7.05  & 0.09  \\
\cline{2-4}
& \textit{TriFusion-AE} & \textbf{0.58 } &\textbf{ 0.011} \\
\hline
\end{tabular}
\end{table}

\paragraph{Analysis} Table \ref{tab-sens} shows that \textit{TriFusion-AE} consistently achieves substantially lower sensitivity values for both MSE and PSNR across all three perturbation types. In particular, under adversarial attacks such as FGSM and PGD, the sensitivity of \textit{TriFusion-AE} is reduced by more than an order of magnitude compared to CNN AE (e.g., $1.23$ vs. $27.39$ for FGSM, and $0.58$ vs. $7.05$ for PGD). Even under stochastic noise, which is less structured than adversarial perturbations, \textit{TriFusion-AE} maintains lower sensitivity. This indicates that multimodal fusion not only enhances robustness against targeted adversarial strategies but also improves resilience to random perturbations.

\section{Discussion}

Across all three robustness evaluations, i.e, stochastic noise, FGSM and PGD, \textit{TriFusion-AE} consistently outperforms CNN AE. Under stochastic perturbations, CNN AE shows rapidly increasing reconstruction error, while \textit{TriFusion-AE} maintains stable MSE and PSNR, reflecting noise-tolerant representations. With FGSM, CNN AE suffers sharp degradation (PSNR $<20$ dB), whereas \textit{TriFusion-AE} sustains higher quality across budgets. Most notably, under the stronger PGD attack, CNN AE collapses, while \textit{TriFusion-AE} preserves fidelity with PSNR remaining above $20$ dB. We also observe a consistently lower sensitivity of \textit{TriFusion-AE} towards stochastic noise intensity and adversarial perturbation, indicating strong robustness of multimodal fusion models. Overall, these results demonstrate that \textit{TriFusion-AE} delivers robustness against both random and adversarial perturbations, with its transformer-based fusion design mitigating error propagation and promoting generalizable features for real-world reconstruction.

\section{Limitations}
While our study demonstrates the benefits of triple-modality fusion in enhancing reconstruction fidelity and adversarial robustness, several limitations remain. First, our evaluation is restricted to controlled benchmarks and simpler model architectures, and does not fully capture real-world sensor miscalibrations, occlusions, or domain shifts [16]. Second, the reliance on a shared autoencoder bottleneck may inadvertently under-represent modality-specific nuances, leading to information loss when one modality is highly degraded. Third, although we study common adversarial perturbations (stochastic noise, FGSM, PGD), the defense generality against more sophisticated attacks remains an open question [17]. Finally, the additional training and inference costs associated with triple-modality encoding could limit deployment in latency-critical applications such as autonomous driving. Addressing these challenges requires future work on scalable architectures, robustness to modality dropout, and evaluation under diverse real-world settings.

\section{Conclusion}

In this work, we investigated the robustness of multimodal fusion architectures under diverse perturbation regimes, including stochastic noise, FGSM, and PGD attacks. Our results demonstrate that visually grounded fusion, when guided by auxiliary textual cues and depth modality, significantly enhances resilience compared to unimodal baselines. Beyond reducing reconstruction errors, multimodal integration enables models to preserve semantic fidelity under adversarial attacks, highlighting the critical role of text and multi-view images in constraining and regularizing 3D visual representations of point clouds.
From a societal perspective, robustness of multimodal systems has profound implications for safety-critical applications such as autonomous driving, robotics, and healthcare. For instance, a self-driving vehicle must maintain reliable perception despite sensor corruption, while a medical imaging system must remain resilient against adversarial perturbations that could compromise diagnoses. Our findings underscore the broader importance of multimodal grounding not only for performance in standard regimes, but also for trustworthy deployment in real-world, adversarially challenging conditions.
Looking ahead, future directions include extending these insights to large-scale datasets, exploring fine-grained cross-attention mechanisms, and developing adaptive fusion strategies that dynamically weigh modalities depending on environmental reliability. More broadly, advancing interpretability alongside robustness of multimodal learning will be essential for building AI systems that are not only accurate, but also transparent, reliable, and safe for real-world decision-making.

\newpage
\section*{References}

{
\small
[1] Hanieh Naderi and Ivan V. Bajić. 2023. Adversarial Attacks and Defenses on 
3D Point Cloud Classification: A Survey. arXiv:2307.00309 [cs.CV]. 
https://arxiv.org/abs/2307.00309

[2] Caesar, H., Bankiti, V., Lang, A., Vora, S., Liong, V., Xu, Q., Krishnan, A., Pan, Y., Baldan, G., \& Beĳbom, O. (2019). nuScenes: A multimodal dataset for autonomous driving. arXiv e-prints, arXiv:1903.11027.

[3] Zhijian Liu, Haotian Tang, Alexander Amini, Xinyu Yang, Huizi Mao, Daniela Rus, and Song Han. 2022. BEVFusion: Multi-Task Multi-Sensor Fusion with Unified Bird’s-Eye View Representation. arXiv e-prints (May 2022), arXiv:2205.13542. https://doi.org/10.48550/arXiv.2205.13542

[4] Yingwei Li, Adams Wei Yu, Tianjian Meng, Ben Caine, Jiquan Ngiam, Daiyi Peng, Junyang Shen, Bo Wu, Yifeng Lu, Denny Zhou, Quoc V. Le, Alan Yuille, and Mingxing Tan. 2022. DeepFusion: LiDAR-Camera Deep Fusion for Multi-Modal 3D Object Detection. (2022). Retrieved from https://arxiv.org/abs/2203.08195

[5] Yichen Xie, Chenfeng Xu, Marie-Julie Rakotosaona, Patrick Rim, Federico Tombari, Kurt Keutzer, Masayoshi Tomizuka, and Wei Zhan. 2023. SparseFusion: Fusing Multi-Modal Sparse Representations for Multi-Sensor 3D Object Detection. (2023). Retrieved from https://arxiv.org/abs/2304.14340

[6] Mingqian Ji, Jian Yang, and Shanshan Zhang. 2025. DepthFusion: Depth-Aware Hybrid Feature Fusion for LiDAR-Camera 3D Object Detection. (2025). Retrieved from https://arxiv.org/abs/2505.07398

[7] Daniel Bogdoll, Yitian Yang, Tim Joseph, Melih Yazgan, and J. Marius Zollner. 2025. MUVO: A Multimodal Generative World Model for Autonomous Driving with Geometric Representations. 2025 IEEE Intelligent Vehicles Symposium (IV) (June 2025), 2243–2250. https://doi.org/10.1109/iv64158.2025.11097718

[8] Jian Zou, Tianyu Huang, Guanglei Yang, Zhenhua Guo, Tao Luo, Chun-Mei Feng, and Wangmeng Zuo. 2024. UniM$^2$AE: Multi-modal Masked Autoencoders with Unified 3D Representation for 3D Perception in Autonomous Driving. (2024). Retrieved from https://arxiv.org/abs/2308.10421

[9] Pei Liu, Haipeng Liu, Haichao Liu, Xin Liu, Jinxin Ni, and Jun Ma. 2025. VLM-E2E: Enhancing End-to-End Autonomous Driving with Multimodal Driver Attention Fusion. arXiv preprint arXiv:2502.18042. https://doi.org/10.48550/arXiv.2502.18042

[10] Ashish Vaswani, Noam Shazeer, Niki Parmar, Jakob Uszkoreit, Llion Jones, Aidan N. Gomez, Łukasz Kaiser, and Illia Polosukhin. 2017. Attention Is All You Need. In Advances in Neural Information Processing Systems (NeurIPS '17), Vol. 30. Curran Associates, Inc. 

[11] Ian J. Goodfellow, Jonathon Shlens, and Christian Szegedy. 2015. Explaining and Harnessing Adversarial Examples. arXiv preprint arXiv:1412.6572.

[12] Nicolas Papernot, Fartash Faghri, Nicholas Carlini, Ian Goodfellow, Reuben Feinman, Alexey Kurakin, Ryan Sheatsley, Abhibhav Garg, Yen-Chen Lin, Scott Cross, Cihang Xie, Yash Sharma, Tom Brown, Aurélien Raffel, Zahra C. K. Li, Milad Nasr, Mohammad Amin Ghasemi, Jiajun Lu, Huan Zhang, Tianhang Zheng, Rujun Long, others, and Patrick McDaniel. 2018. CleverHans v2.1.0: an adversarial machine learning library. arXiv preprint arXiv:1610.00768.

[13] Aleksander Madry, Aleksandar Makelov, Ludwig Schmidt, Dimitris Tsipras, and Adrian Vladu. 2018. Towards Deep Learning Models Resistant to Adversarial Attacks. In International Conference on Learning Representations (ICLR).

[14] Alec Radford, Jong Wook Kim, Chris Hallacy, Aditya Ramesh, Gabriel Goh, Sandhini Agarwal, Girish Sastry, Amanda Askell, Pamela Mishkin, Jack Clark, Gretchen Krueger, and Ilya Sutskever. 2021. Learning transferable visual models from natural language supervision. In Proceedings of the 38th International Conference on Machine Learning (ICML 2021). PMLR, 8748–8763.

[15] Alexey Dosovitskiy, Lucas Beyer, Alexander Kolesnikov, Dirk Weissenborn, Xiaohua Zhai, Thomas Unterthiner, Mostafa Dehghani, Matthias Minderer, Georg Heigold, Sylvain Gelly, Jakob Uszkoreit, and Neil Houlsby. 2021. An image is worth 16x16 words: Transformers for image recognition at scale. In International Conference on Learning Representations (ICLR 2021).

[16] Wang, Z., Huang, Y., Luo, W., Xie, T., Jing, M., and Zuo, L. 2025. SWAT: Sliding Window Adversarial Training for Gradual Domain Adaptation. arXiv preprint arXiv:2501.19155. Available at: https://arxiv.org/abs/2501.19155

[17] Chahe, A., Wang, C., Jeyapratap, A., Xu, K., and Zhou, L. 2024. Dynamic Adversarial Attacks on Autonomous Driving Systems. arXiv preprint arXiv:2312.06701.

}

\end{document}